# Supervised feature selection with orthogonal regression and feature weighting

Xia Wu, Xueyuan Xu, Jianhong Liu, Hailing Wang, Bin Hu, Feiping Nie*

**Abstract**—Effective features can improve the performance of a model, which can thus help us understand the characteristics and underlying structure of complex data. Previous feature selection methods usually cannot keep more local structure information. To address the defects previously mentioned, we propose a novel supervised orthogonal least square regression model with feature weighting for feature selection. The optimization problem of the objection function can be solved by employing generalized power iteration (GPI) and augmented Lagrangian multiplier (ALM) methods. Experimental results show that the proposed method can more effectively reduce the feature dimensionality and obtain better classification results than traditional feature selection methods. The convergence of our iterative method is proved as well. Consequently, the effectiveness and superiority of the proposed method are verified both theoretically and experimentally.

## I. INTRODUCTION

FEATURE selection is one of the most important research problems in machine learning, devoted to selecting the most effective elements from the original features in order to reduce the overall dimensions of high-dimensional data sets and improve the performance of learning algorithms [1].

In general, there are three types of feature selection methods: filter methods, wrapper methods, and embedded methods [2]. Filter methods implement feature selection before classification and are usually based on a two-step strategy. First, the features are ranked by certain criteria. Second, the features with top scores are selected. There are many classic filter methods, such as the Chi-squared test, information gain (IG) [3], correlation coefficient (CC) scores, maximum relevance minimum redundancy (mRMR) [4], correlation-based feature selection (CFS) [5] and so forth. Wrapper methods generate different subsets of features and then evaluate the subsets under certain classifiers or learning algorithms, for example the Genetic Algorithm (GA) [6]. Embedded methods are similar to filter methods, but they determine the sort of features through training. L1 (LASSO) regularization [7] and decision trees [8] are typical examples of embedded methods.

This work was supported by the General Program of National Natural Science Foundation of China No. 61876021.

X. Wu, X.Y. Xu, J. H. Liu, H. L. Wang, and B. Hu are with the College of Information Science and Technology, Beijing Normal University, Beijing, CN 100875.

F. P. Nie*(the corresponding author, feipingnie@gmail.com) is with the School of Computer Science, OPTIMAL, Northwestern Polytechnical University, Xian, CN 710072.

Feature selection can also be sorted into supervised, semi-supervised, and unsupervised methods based on class label information. Supervised feature selection methods evaluate feature relevance using all class labels, for example Fisher Score [9] and ReliefF [10]. Recently, various semi-supervised methods, with some class labels, have been proposed, for example Chen et al. [11] proposed a semi-supervised feature selection utilizing rescaled linear regression. Without labels, unsupervised feature selection methods compute feature relevance through feature similarity, for example the Laplacian Score [12].

Least square regression is the most common statistical analysis model. It aims to find a projection matrix $W$ and minimizes a sum-of-squares error function [13]. At present, some feature selection methods exist that have been put forward based on least square regression. For example, Sparse LSR was proposed by Nie et al. in [14] and introduced $l_{2,1}$ to sparse the projection matrix $W$ so as to select the effective features. The orthogonal regression model can be considered the least square regression with orthogonal constrains. It can preserve more discrimination information in a subspace, and avoid trivial solutions, compared to the least square regression [15]. The optimization problem of the classical least square regression can be solved easily. However, the objective function with orthogonal constrains is an unbalanced orthogonal procrustes problem which is difficult to obtain an optimal solution.

In this paper we propose a novel supervised feature selection method, named Feature Selection with Orthogonal Regression (FSOR). The proposed method is a technique for feature selection, based on orthogonal regression, which aims to minimize the perpendicular distance from the data points to the fitted function. Unlike other classical orthogonal regression models, we introduce the feature weighting information in our model. The new scale factors can express the ranking or proportion of all features with the aim to minimize the perpendicular distance from the data points to the fitted function. In other words, they are used to evaluate the importance of features. The feature subsets are then formed by selected the features with the top rankings. Moreover, motivated by the previous study [16], the generalized power iteration (GPI) method was employed in our work to solve the regression matrix $W$, and an effective iterative algorithm was derived to minimize the objective function. To assess the reliability, the proposed method FSOR was compared with eight other state-of-the-art supervised feature selection methods, including ReliefF, CC, IG, trace ratio criterion (TRC) [17], robust feature selection (RFS) [14], conditional mutual information maximization criterion (CMIM) [18], Fisher and mRMR. The experiment results prove the convergence of our method and show the superiority on various data sets when compared with five feature selection methods.

The rest of the paper is organized as the follows. Section II gives the notations and definitions of the norms employed in this paper. In Section III, we give a detailed introduction of the proposed FSOR method. Next, in Section IV, the experiment results and discussions are presented. Finally, a conclusion is given in Section V.

## II. NOTATIONS AND DEFINITIONS

We summarize the notations and definitions of the norms that are used in this paper. $I_n$ denotes an $n \times n$ identity matrix. $\mathbf{1}_n = (1,1,...1)^T \in R^{n \times 1}$. For any matrix $M$, the Frobenius norm is defined as $\|M\|_F^2 = Tr(M^T M)$.

## III. PROPOSED METHOD

In [19], the orthogonal least square regression (OLSR) can be written as:

$$\min_{W,b} \|W^T X + b\mathbf{1}_n^T - Y\|_F^2 \quad \text{s.t.} \quad W^T W = I_k \quad (1)$$

where the data matrix $X \in R^{d \times n}$, the label matrix $Y \in R^{k \times n}$, the regression matrix $W \in R^{d \times k}$ with orthogonal constrain $W^T W = I_k$, and $b \in R^{k \times 1}$ is the bias vector. $d$, $n$ and $k$ represent the number of features, samples and categories, respectively.

To express the ranking or proportion of all features, we introduce feature weighting information in our model. Based on the OLSR and feature weighting, we propose a new supervised feature selection method by solving the optimization problem, this can be written as:

$$\min_{W,b,\theta} \|W^T \Theta X + b\mathbf{1}_n^T - Y\|_F^2 \quad (2)$$

$$\text{s.t.} \quad W^T W = I_k, \theta^T \mathbf{1}_d = 1, \theta \geq 0$$

Where the diagonal matrix, $\Theta \in R^{d \times d}$ with $\theta^T \mathbf{1}_d = 1$ and $\theta \geq 0$, measures the importance of features. Due to the extreme value condition w.r.t. $b$, we can derive that:

$$\frac{\partial \|W^T \Theta X + b\mathbf{1}_n^T - Y\|_F^2}{\partial b} = 0 \quad (3)$$

Where $b$ can be computed as

$$b = \frac{1}{n}(Y\mathbf{1}_n - W^T \Theta X \mathbf{1}_n) \quad (4)$$

Substituting Eq. (4) into Eq. (2), Eq. (2) can be simplified to the following form:

$$\min_{W,\theta} \|W^T \Theta XH - YH\|_F^2 \quad (5)$$

$$\text{s.t.} \quad W^T W = I_k, \theta^T \mathbf{1}_d = 1, \theta \geq 0$$

where $H = I_n - \frac{1}{n}\mathbf{1}_n\mathbf{1}_n^T$

Hence, the problem can be converted to Eq. (5). In the following, we propose a new effective method to solve the problem (5).

### A. Update $W$ with $\Theta$ fixed

When fixing $\Theta$, we can deduce the following formula

$$\min_{W^T W = I_k} Tr(W^T AW - 2W^T B) \quad (6)$$

in which: $\begin{cases} A = \Theta X H X^T \Theta^T \\ B = \Theta X H Y^T \end{cases}$

Eq. (6) has the same form as the quadratic problem on the Stiefel manifold (QPSM) [18], where $W \in R^{d \times k}$, $B \in R^{d \times k}$ and symmetric matrix $A \in R^{d \times d}$. Nie et al. propose a novel generalized power iteration (GPI) method to solve the QPSM [20], and whereby the efficiency of the GPI method is verified both theoretically and empirically. The algorithm of the GPI is described in Algorithm 1.

Thus, we can update $W$ when $\Theta$ is fixed using the Algorithm 1.

---

**Algorithm 1** Generalized power iteration (GPI) method

1. **Input**: the symmetric matrix $A \in R^{d \times d}$ and matrix $B \in R^{d \times k}$.
2. **Output**: the matrix $W \in R^{d \times k}$.
3. **Initialize** the random $W \in R^{d \times k}$ and parameter $\alpha$ such as $\tilde{A} = \alpha I_d - A \in R^{d \times d}$ is a positive definite matrix.
4. **Repeat**
5.    Update $M \leftarrow 2\tilde{A}W + 2B$.
6.    Calculate $USV^T = M$ via the compact SVD method of $M$.
7.    Update $W \leftarrow UV^T$.
8. **until** *converges*.

---

### B. Update $\Theta$ with $W$ fixed

When $W$ is fixed, the problem (5) becomes

$$\min_{W,b,\theta}\left[Tr(\Theta X H X^T \Theta W W^T) - Tr(2\Theta X H Y^T W^T)\right] \quad (7)$$

$$\text{s.t.} \quad W^T W = I_k, \theta^T \mathbf{1}_d = 1, \theta \geq 0$$

Lemma 1. If $S$ is diagonal, then $Tr(SASB) = s^T(A^T \circ B)s$

Proof:
$Tr(SASB) = s^T diag(ASB) = s^T vec\{a_i^T Sb_i\} = s^T vec\{(a_i \circ b_i)^T s\}$
$= s^T(A^T \circ B)^T s = s^T(A^T \circ B)s$

By using the lemma 1, the problem (7) becomes

$$\min_{W,b,\theta}\left[\theta^T\left[(XHX^T)^T \circ (WW^T)\right]\theta - \theta^T b\right] \quad (8)$$

$$\text{s.t.} \quad W^T W = I_k, \theta^T \mathbf{1}_d = 1, \theta \geq 0$$

The problem (8) can be written as the following form

$$\min_{\theta^T \mathbf{1}_d = 1, \theta \geq 0} \theta^T A \theta - \theta^T b \quad (9)$$

in which: $\begin{cases} A = (XH^T X^T) \circ (WW^T) \\ b = diag(2XHY^T W^T) \end{cases}$

Therefore, problem (5) is converted into a solution to the problem (9). We use the augmented Lagrangian multiplier (ALM) method to solve the constrained minimization problem

3and decompose the problem into multiple subproblems [21, 22]. The ALM method is introduced to solve the following constrained minimization problem:

$$\min_{\varphi(X)=0} f(X) \quad (10)$$

The solution is described in Algorithm 2.

**Algorithm 2** The augmented Lagrangian multiplier (ALM) method
1. **Set** $1<\rho<2$, **initialize** $\mu>0$, $\lambda$.
2. **Output**: $X$.
3. **repeat**
4.    Update $X$ by $\min_X f(X) + \frac{\mu}{2}\left\|\varphi(X) + \frac{1}{\mu}\lambda\right\|_F^2$.
5.    Update $\lambda$ by $\lambda = \lambda + \mu\varphi(X)$.
6.    Update $\mu$ by $\mu = \rho\mu$.
7. **Until** *convergence*.

We rewrite the problem (9) as the following:

$$\min_{\theta^T \mathbf{1}_d=1, v\geq 0, v=\theta} \theta^T A \theta - \theta^T b \quad (11)$$

According to Algorithm 2, the augmented Lagrangian function of (9) is defined as:

$$L(\theta, v, \mu, \lambda_1, \lambda_2) = \theta^T A\theta - \theta^T b + \frac{\mu}{2}\left\|\theta - v + \frac{1}{\mu}\lambda_1\right\|_F^2$$
$$+ \frac{\mu}{2}\left(\theta^T \mathbf{1}_d - 1 + \frac{1}{\mu}\lambda_2\right)^2 \quad (12)$$
$$\text{s.t. } v \geq 0$$

where $v$ and $\lambda_1$ are column vectors, $\mu$ is the Lagrangian multipliers. When fixing $v$, problem (12) can be equivalently rewritten as:

$$\min_\theta \frac{1}{2}\theta^T E \theta - \theta^T f \quad (13)$$

in which: $\begin{cases} E = 2A + \mu I_d + \mu\mathbf{1}_d\mathbf{1}_d^T \\ f = \mu v + \mu\mathbf{1}_d - \lambda_2\mathbf{1}_d - \lambda_1 + b \end{cases}$

It can be easily seen that $\theta = E^{-1}f$.

In the same way, when fixing $\theta$, problem (12) can be rewritten as:

$$\min_{v\geq 0}\left\|v - \left(\theta + \frac{1}{\mu}\lambda_1\right)\right\|^2 \quad (14)$$

It can be verified that the optimal solution of $v$ is

$$\hat{v} = pos(\theta + \frac{1}{\mu}\lambda_1) \quad (15)$$

$pos(t)$ is a function which assigns 0 to each negative element of $t$.

To sum up, the detailed algorithm for solving the problem (9) is described in Algorithm 3.

**Algorithm 3** Algorithm to solve the problem (9)
1. **Initialize** $\rho>1$, $\theta_i = \frac{1}{d}(1\leq i \leq d)$, $v=\theta$, $\lambda_2=0$, $\mu>0$, $\lambda_1 = (0,0,...0)^T \in R^{d\times 1}$.
2. **Output**: $\theta$
3. **Repeat**
4.    Update $E$ by $E = 2A + \mu I_d + \mu\mathbf{1}_d\mathbf{1}_d^T$
5.    Update $f$ by $f = \mu v + \mu\mathbf{1}_d - \lambda_2\mathbf{1}_d - \lambda_1 + b$
6.    Update $\theta$ by $\theta = E^{-1}f$
7.    Update $v$ by $\hat{v} = pos(\theta + \frac{1}{\mu}\lambda_1)$
8.    Update $\lambda_1$ by $\lambda_1 = \lambda_1 + \mu(\theta - v)$
9.    Update $\lambda_2$ by $\lambda_2 = \lambda_2 + \mu(\theta^T\mathbf{1}_d - 1)$
10.   Update $\mu$ by $\mu = \rho\mu$
11. **Until** *convergence*.

The detailed algorithm for solving the problem (2), named Feature Selection with Orthogonal Regression (FSOR), is summarized in Algorithm 4. In this algorithm, the regression matrix $W$ and the diagonal matrix $\Theta$ are alternately updated until convergence.

**Algorithm 4** Feature Selection with Orthogonal Regression (FSOR) method
1. **Input**: the data matrix $X \in R^{d\times n}$, the label matrix $Y \in R^{k\times n}$
2. **Output**: the regression matrix $W \in R^{d\times k}$, the diagonal matrix $\Theta \in R^{d\times d}$.
3. **Initialize** $\Theta \in R^{d\times d}$ satisfying $\theta^T\mathbf{1}_d = 1$ and $\theta \geq 0$.
$$H = I_n - \frac{1}{n}\mathbf{1}_n\mathbf{1}_n^T$$
4. **repeat**
5.    Update $W$ via Algorithm 1.
6.    Update $\Theta$ via Algorithm 3.
7. **Until** *convergence*.

## IV. EXPERIMENTAL RESULTS AND DISCUSSION

To verify the correctness and reliability of our algorithm, we chose eight Benchmark datasets from Feiping Nie's page[1] as the experimental data. The detailed information regrading these datasets is summarized in Table I.

TABLE I
DESCRIPTION OF 8 BENCHMARK DATASETS

| Datasets | #of samples | #Features | #Classes |
|---|---|---|---|
| Vehicle | 846 | 18 | 4 |
| Segment | 2310 | 19 | 7 |
| Chess | 3196 | 36 | 2 |
| Control | 600 | 60 | 6 |
| Uspst | 2007 | 256 | 9 |
| Binalpha | 1404 | 320 | 36 |
| Corel_5k | 5000 | 423 | 50 |
| Yeast | 1484 | 1470 | 10 |

---

[1] http://www.escience.cn/system/file?fileId=82035

In the experiment, we compared the FSOR algorithm with eight other state-of-the-art feature selection algorithms, including ReliefF, CC, IG, CMIM, fisher, mRMR, TRC and RFS. In view of the above eight datasets, we employed four classifiers to classify the datasets after feature selection, such as support vector machine (SVM) with linear kernels, SVM with RBF kernels, Random Forest (RF) and k-nearest neighbor (KNN). For the selected features, we used a random 70% of the data to act as the training sets to train the best classifier model, then we test the model using the remaining 30% testing sets. We calculated the results 100 times and took an average value as the final classification accuracy. It should be noted that we use the LIBSVM toolbox to employ the SVM classifier.

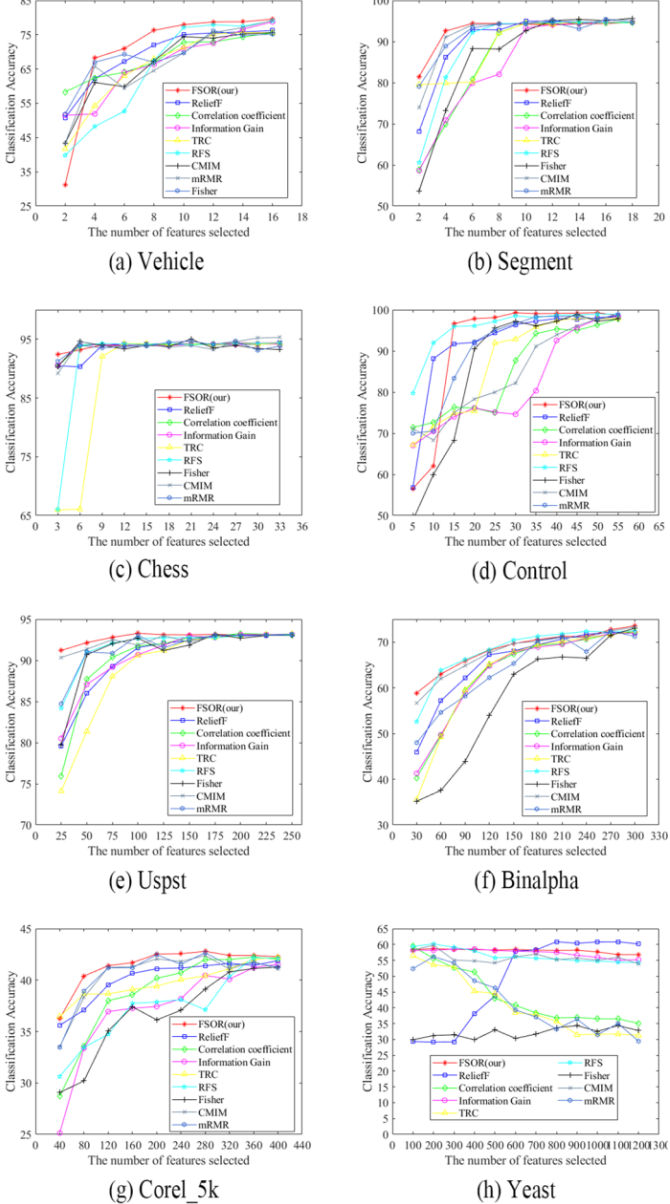

Fig. 1. Magnetization Classification accuracy vs. dimension.

The varied circumstances of classification accuracy of SVM with linear kernels under different feature selection methods for each dataset are described in Fig. 1. The comparisons of average classification accuracy for all the feature set sizes, and deviation using FSOR, and the other eight feature selection methods for the eight benchmark datasets are performed in Table II. We can see that the FSOR algorithm performs well in most datasets. The accuracy gained by employing the FSOR method is the highest and the deviation is the lowest, excluding the control dataset. Furthermore, the FSOR method also works well on the Control dataset and only performes worse than TRC method.

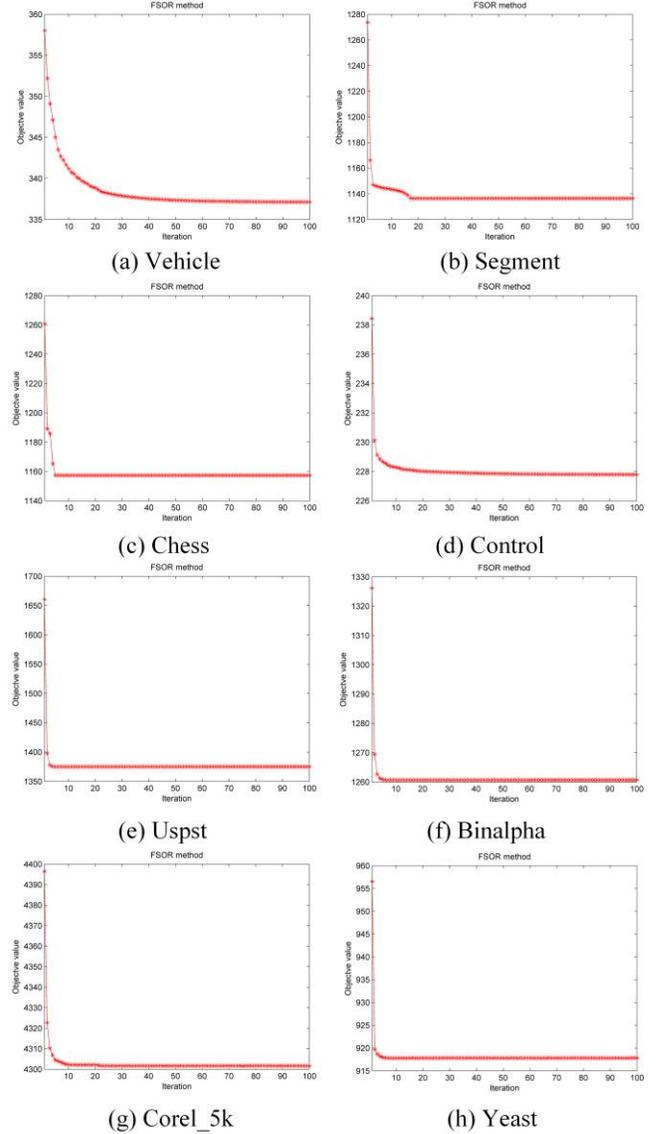

Fig. 2. The convergence of FSOR algorithm.

As shown in Fig. 1(b, c, d, e, g, h), the recognition accuracy of the proposed FSOR method first keeps growing as the feature number size increases, and then falls down slightly. This can be attributed to the fact that, initially, more features can provide more information to distinguish samples belonging to different classes, but, as the number of features continues to increase excessively, the features with noisy information might be added into the selected feature subset and thus reduce the recognition accuracy of the FSOR method. Compared with other feature selection methods, the recognition rate of the proposed method increases to the highest value significantly faster, and then maintains more stable. This is because the

TABLE II
THE COMPARISONS OF AVERAGE CLASSIFICATION ACCURACY (%) OF 9 FEATURE SELECTION METHODS AND FOUR CLASSIFIERS ARE PERFORMED ON 8 BENCHMARK DATASETS. HERE, THE "*" INDICATES THE DIFFERENCE BETWEEN THE RESULTS OF FSOR AND THOSE OF THE CORRESPONDING ALGORITHM (EXCLUDING FSOR) IS SIGNIFICANT BY T-TEST, I.E., THE P-VALUE OF T-TEST IS LESS THAN 0.05.

| Dataset | Vehicle | | | | Segment | | | |
|---|---|---|---|---|---|---|---|---|
| Accuracy | SVM (linear) | SVM (rbf) | KNN | RF | SVM (linear) | SVM (rbf) | KNN | RF |
| ReliefF | 69.33 ± 0.71 | 72.79 ± 0.28 | 68.16 ± 0.07 | 72.08 ± 0.36 | 90.46 ± 0.69 | 94.47 ± 0.22 | 93.22 ± 0.30 | 95.37 ± 0.24 |
| TRC | 65.57 ± 1.30 | 70.13 ± 0.32 | 60.88 ± 0.19 | 67.13 ± 0.69 | 89.39 ± 0.46 | 94.37 ± 0.14 | 93.63 ± 0.16 | 94.15 ± 0.20 |
| RFS | 64.96 ± 2.17 | 71.46 ± 1.03 | 64.42 ± 0.67 | 69.64 ± 0.83 | 89.07 ± 1.18 | 94.08 ± 0.37 | 92.53 ± 0.38 | 94.47 ± 0.46 |
| Fisher | 66.34 ± 1.27 | 66.29 ± 0.47 | 64.17 ± 0.19 | 66.88 ± 0.78 | 86.36 ± 2.01 | 92.16 ± 1.78 | 90.11 ± 1.57 | 93.23 ± 1.27 |
| CMIM | 67.27 ± 1.48 | 66.34 ± 0.46 | 63.83 ± 0.12 | 69.19 ± 0.80 | 91.97 ± 0.47 | 95.03 ± 0.16 | 94.56 ± 0.08 | 96.01 ± **0.11** |
| mRMR | 69.22 ± 0.68 | 68.55 ± 0.17 | 60.38 ± 0.10 | 71.51 ± 0.40 | 91.97 ± 0.26 | 94.94 ± 0.20 | 94.79 ± 0.11 | 95.99 ± 0.12 |
| CC | 68.40 ± **0.36** | 67.42 ± 0.19 | 62.20 ± 0.04 | 70.13 ± 0.37 | 86.03 ± 1.58 | 91.81 ± 0.80 | 91.01 ± 0.89 | 91.74 ± 0.99 |
| IG | 66.56 ± 0.94 | 65.99 ± 0.25 | 61.71 ± 0.25 | 68.01 ± 0.51 | 84.74 ± 1.51 | 93.07 ± 0.27 | 91.97 ± 0.27 | 92.91 ± 0.43 |
| FSOR(our) | **69.37** ± 2.32 | **73.08*** ± **0.07** | **68.18** ± **0.03** | **72.29** ± **0.30** | **92.16** ± **0.22** | **95.54** ± **0.12** | **95.34 *** ± **0.05** | **96.17** ± **0.11** |

| Dataset | Chess | | | | Control | | | |
|---|---|---|---|---|---|---|---|---|
| Accuracy | SVM (linear) | SVM (rbf) | KNN | RF | SVM (linear) | SVM (rbf) | KNN | RF |
| ReliefF | 93.48 ± 0.02 | 95.88 ± 0.11 | 94.66 ± 0.04 | 94.96 ± 0.06 | 91.45 ± 1.33 | 92.22 ± 1.26 | 94.40 ± 0.78 | 91.18 ± 0.81 |
| TRC | 88.88 ± 1.16 | 94.94 ± 0.07 | 94.28 ± 0.03 | 95.17 ± **0.03** | 87.33 ± 1.38 | 91.72 ± **0.64** | 92.42 ± 0.44 | 91.31 ± 0.43 |
| RFS | 91.63 ± 0.65 | 92.87 ± 0.06 | 91.51 ± 0.85 | 92.62 ± 0.89 | **94.55** ± **0.54** | 92.07 ± 1.36 | **96.72** ± **0.22** | **95.11** ± **0.26** |
| Fisher | 93.72 ± 0.02 | 95.39 ± 0.75 | 94.16 ± 0.03 | 95.36 ± 0.05 | 86.16 ± 3.26 | 87.17 ± 3.57 | 86.16 ± 3.55 | 86.26 ± 3.67 |
| CMIM | 93.76 ± 0.03 | 95.97 ± 0.05 | 94.39 ± **0.02** | 95.55 ± 0.04 | 84.75 ± 1.23 | 92.12 ± 1.37 | 94.14 ± 0.69 | 90.66 ± 0.55 |
| mRMR | 93.74 ± 0.01 | 95.56 ± 0.08 | 94.15 ± 0.03 | 95.54 ± 0.08 | 91.06 ± 1.27 | 92.22 ± 1.25 | 94.30 ± 0.76 | 90.51 ± 0.67 |
| CC | 93.47 ± 0.02 | 95.12 ± **0.04** | 94.15 ± 0.03 | 94.89 ± 0.05 | 85.25 ± 1.08 | 92.32 ± 0.66 | 92.27 ± 0.60 | 90.20 ± 0.48 |
| IG | 93.75 ± 0.01 | 96.01 ± 0.06 | 94.51 ± 0.03 | 95.58 ± **0.03** | 82.10 ± 1.25 | 92.47 ± 0.82 | 92.02 ± 0.46 | 90.51 ± 0.52 |
| FSOR(our) | **93.84** ± **0.01** | **96.13** ± 0.10 | **94.92** ± **0.02** | **96.25*** ± **0.04** | 91.48* ± 2.32 | **92.88** ± 0.96 | 94.69 ± 0.67 | 91.36* ± 0.53 |

| Dataset | Uspst | | | | Binalpha | | | |
|---|---|---|---|---|---|---|---|---|
| Accuracy | SVM (linear) | SVM (rbf) | KNN | RF | SVM (linear) | SVM (rbf) | KNN | RF |
| ReliefF | 90.30 ± 0.17 | 94.49 ± 0.11 | 91.88 ± 0.24 | 90.53 ± 0.16 | 65.66 ± 0.64 | 66.46 ± 0.88 | 60.24 ± 0.78 | 63.18 ± 1.00 |
| TRC | 89.00 ± 0.37 | 92.91 ± 0.20 | 91.31 ± 0.28 | 91.01 ± 0.17 | 63.19 ± 1.30 | 65.65 ± 1.35 | 59.92 ± 1.32 | 62.07 ± 0.75 |
| RFS | 91.75 ± 0.07 | 94.37 ± 0.05 | 92.66 ± 0.14 | 91.54 ± 0.12 | 68.15 ± 0.34 | 71.76 ± 0.31 | 64.44 ± 0.29 | 64.70 ± 0.20 |
| Fisher | 91.10 ± 0.17 | 93.64 ± 0.17 | 90.83 ± 0.24 | 90.03 ± 0.20 | 54.96 ± 3.99 | 58.72 ± 3.26 | 51.73 ± 2.58 | 54.32 ± 2.95 |
| CMIM | 92.57 ± 0.01 | 95.32 ± **0.01** | 92.72 ± 0.20 | 92.87 ± **0.01** | 68.22 ± 0.31 | 70.83 ± 0.22 | 65.04 ± 0.16 | 67.67 ± 0.06 |
| mRMR | 91.54 ± 0.07 | 94.72 ± 0.06 | 93.02 ± 0.08 | 92.11 ± 0.05 | 64.35 ± 0.72 | 67.08 ± 0.78 | 62.64 ± 0.70 | 62.95 ± 0.73 |
| CC | 90.30 ± 0.26 | 93.11 ± 0.42 | 90.96 ± 0.37 | 90.10 ± 0.29 | 63.51 ± 1.03 | 64.58 ± 1.26 | 59.57 ± 1.35 | 62.07 ± 1.04 |
| IG | 90.46 ± 0.15 | 93.27 ± 0.16 | 90.42 ± 0.27 | 90.32 ± 0.13 | 63.55 ± 0.99 | 65.63 ± 1.21 | 60.12 ± 1.44 | 61.90 ± 0.87 |
| FSOR(our) | **92.83** ± **0.01** | **95.54** ± **0.01** | **93.13** ± 0.05 | **93.01** ± **0.01** | **68.25** ± **0.17** | **72.07** ± **0.16** | **65.46 *** ± **0.15** | **67.83 *** ± **0.05** |

| Dataset | Corel_5k | | | | Yeast | | | |
|---|---|---|---|---|---|---|---|---|
| Accuracy | SVM (linear) | SVM (rbf) | KNN | RF | SVM (linear) | SVM (rbf) | KNN | RF |
| ReliefF | 40.15 ± 0.04 | 41.08 ± 0.08 | 29.96 ± 0.09 | 41.31 ± 0.03 | 49.08 ± 1.80 | 43.60 ± 1.57 | 40.24 ± 2.30 | 41.91 ± 1.38 |
| TRC | 39.81 ± 0.03 | 38.81 ± 0.05 | 27.23 ± 0.09 | 39.06 ± 0.01 | 40.91 ± 0.81 | 57.83 ± 0.16 | 42.32 ± 0.93 | 57.77 ± 0.06 |
| RFS | 37.43 ± 0.12 | 39.73 ± 0.07 | 28.35 ± 0.08 | 36.92 ± 0.11 | 56.50 ± 0.04 | 60.41 ± 0.21 | 55.00 ± 0.06 | 58.26 ± 0.04 |
| Fisher | 36.73 ± 0.19 | 37.87 ± 0.16 | 26.10 ± 0.19 | 34.89 ± 0.18 | 32.12 ± 0.02 | 48.50 ± 0.86 | 31.40 ± **0.02** | 34.29 ± 1.07 |
| CMIM | 40.52 ± 0.07 | 42.73 ± 0.03 | 31.90 ± **0.02** | 41.82 ± 0.02 | 55.41 ± 0.10 | 61.87 ± 0.19 | 54.70 ± 0.09 | 58.73 ± 0.04 |
| mRMR | 40.58 ± 0.07 | 42.73 ± 0.03 | 31.63 ± 0.03 | 42.17 ± 0.02 | 41.63 ± 0.88 | 57.68 ± 0.15 | 42.23 ± 0.75 | 59.95 ± 0.07 |
| CC | 38.80 ± 0.18 | 39.49 ± 0.22 | 27.63 ± 0.22 | 37.39 ± 0.17 | 43.58 ± 0.69 | 61.65 ± 0.29 | 43.69 ± 0.60 | 57.02 ± **0.03** |
| IG | 37.17 ± 0.22 | 38.18 ± 0.29 | 27.64 ± 0.19 | 37.50 ± 0.19 | 57.30 ± 0.02 | 57.73 ± **0.04** | 56.21 ± 0.10 | 59.61 ± 0.05 |
| FSOR(our) | **40.60** ± **0.02** | **43.03** ± **0.02** | **32.53** ± **0.02** | **42.33** ± **0.01** | **58.04** ± **0.01** | **63.35*** ± **0.04** | 56.98 ± 0.02 | **60.46** ± 0.03 |

TABLE III
THE COMPARISONS OF SENSITIVITY AND SPECIFICITY (%) OF 9 FEATURE SELECTION METHODS AND FOUR CLASSIFIERS ARE PERFORMED ON YEAST DATASETS.

| Yeast dataset | Sensitivity | | | | Specitivity | | | |
|---|---|---|---|---|---|---|---|---|
| (%) | SVM (linear) | SVM (rbf) | KNN | RF | SVM (linear) | SVM (rbf) | KNN | RF |
| ReliefF | 29.11 | 29.56 | 26.67 | 22.74 | 91.95 | 92.18 | 92.06 | 91.75 |
| TRC | 17.39 | 27.90 | 17.80 | 44.48 | 91.33 | 94.34 | 91.55 | 93.94 |
| RFS | 35.26 | 38.41 | 36.58 | 42.83 | 93.93 | 94.55 | 93.89 | 94.29 |
| Fisher | 10.15 | 16.18 | 10.24 | 14.69 | 89.93 | 92.56 | 89.84 | 90.52 |
| CMIM | 40.41 | 46.65 | 38.69 | 43.95 | 93.85 | 94.80 | 93.83 | 94.30 |
| mRMR | 17.13 | 28.04 | 16.90 | **46.90** | 91.64 | 94.31 | 91.64 | 94.31 |
| CC | 37.49 | 45.59 | 38.31 | 42.06 | 92.17 | 94.71 | 92.15 | 94.02 |
| IG | 48.05 | 48.18 | 44.14 | 44.96 | 94.14 | 94.15 | **94.20** | 94.46 |
| FSOR(our) | **51.83** | **48.81** | **45.53** | 45.90 | **94.19** | **95.03** | 93.96 | **94.57** |

FSOR method introduces orthogonal constrains to limit the projection matrix W, which can preserve more discrimination information in subspace and avoid redundant and noisy information. Additionally, compared with the other popular algorithms, the results in Table II show the best recognition rate, collected during implementation of the proposed method, and hence the proposed method can be suggested to be more robust as its performance doesn't depend on a particular classifier.

To further evaluate the multi-class recognition ability, we also computed the specificity and sensitivity of the classification results. Taking the Yeast dataset as an example, Table III shows the average specificity and sensitivity of all the feature set sizes. Significantly, the FSOR algorithm is better than other algorithms overall.

In addition, to prove effectiveness and stability, we tracked the changes in the FSOR's objective function under different datasets. Here the number of iterations was set to 100. As illustrated in Fig. 2, we were able to identify that the objective function values decline and converge to a local minimum, step-by-step, with an increase in the number of iterations.

TABLE IV
COMPUTATIONAL COMPLEXITY OF FEATURE SELECTION METHODS

| Method | Computational Complexity |
|---|---|
| CC | $\mathcal{O}(dn)$ |
| IG | $\mathcal{O}(dn)$ |
| ReliefF | $\mathcal{O}(dn)$ |
| mRMR | $\mathcal{O}(dmn)$ |
| CMIM | $\mathcal{O}(dmn)$ |
| Fisher | $\mathcal{O}(dn)$ |
| TRC | $\mathcal{O}(dn^2 + d^2n)$ |
| RFS | $\mathcal{O}(d^3 + d^2n + dn^2 + n^3 + dkn)$ |
| FSOR (our) | $\mathcal{O}(dkn)$ |

Computational complexity is a key indicator regarding the performance of a certain method. In this section, we analyze the computational complexity of the proposed FSOR method, based on the number of multiplications of the matrix operations. The computational complexity of the FSOR algorithm can be mainly attributed to the calculation of the matrix W of the GPI algorithm. To reduce the computational complexity of the GPI algorithm, we choose to compute the matrix W directly instead of calculating A and then multiplying by W. The calculation of AW requires computational complexity to be determined of the order of $\mathcal{O}(ndk)$. Hence, the order of the computational complexity of FSOR is $\mathcal{O}(ndk)$.

Moreover, the computational complexity of the state-of-the-art supervised feature selection algorithms is provided in Table IV. In Table IV, m is the number of selected features ($1 \leq m < d$). From the results, we could conclude that the computational complexity of FSOR is much less than that of the TRC and RFS methods and depends on d, n and k. When the value of k is small, the computational complexity of FSOR is approximately equal to that of CC, IG, ReliefF and Fisher. However, when the proposed method was implemented with the Binalpha and Corel_5k datasets, the computational time of FSOR was much longer than those of the filter methods. This can be attributed to the value of k being not much smaller than that of n and d.

## V. CONCLUSION

In this paper, we have proposed a novel supervised feature selection algorithm named FSOR. This new method extends the orthogonal least square regression by adding feature weighting, which is used to evaluate the importance of features. Based on the existed GPI algorithm and ALM algorithm, the FSOR decreases the objective value of the model to a local minimum until convergence. Subsequently, we employ the FSOR method on the benchmark datasets and utilize eight feature selection methods as reference. From the experimental results, we reach the projected goal that, overall, the performance of FSOR is superior to the other eight state-of-the-art feature selection methods.

## REFERENCES

[1] M. Dash, H. Liu, "Feature Selection for Classification," *Intelligent data analysis*, vol. 1, no. 4, pp. 131–156, 1997.

[2] Y. Saeys, I. Inza and P. Larrañaga, "A review of feature selection techniques in bioinformatics," *Bioinformatics*, vol. 23, no. 19, pp. 2507-2517, 2007.

[3] L. E. Raileanu and K. Stoffel, "Theoretical Comparison between the Gini Index and Information Gain Criteria," *Annals of Mathematics & Artificial Intelligence*, vol. 41, no. 1, pp. 77–93, 2007.

[4] H. C. Peng, F. H. Long and C. Ding, "Feature Selection Based on Mutual

Information: Criteria of Max-Dependency, Max-Relevance, and Min-Redundancy," *IEEE Transactions on Pattern Analysis & Machine Intelligence,* vol. 27, no. 8, pp. 1226–1234, 2005.

[5] M. A. Hall, L. A. Smith, "Feature Selection for Machine Learning: Comparing a Correlation-Based Filter Approach to the Wrapper," *proc. Twelfth International Florida Artificial Intelligence Research Society Conference*, 1999, pp. 235-239.

[6] J. H. Holland, Adaptation in natural and artificial systems: an introductory analysis with applications to biology, control, and artificial intelligence, MIT Press, 1992.

[7] P. Wei, Q. H. Hu, P. Ma, and X. H. Su, "Robust feature selection based on regularized brownboost loss," *Knowledge-Based Systems*, vol. 54, no. 4, pp. 180–198, 2013.

[8] V. Sugumaran, V. Muralidharan, and K. I. Ramachandran, "Feature selection using Decision Tree and classification through Proximal Support Vector Machine for fault diagnostics of roller bearing", *Mechanical Systems & Signal Processing*, vol. 21, no. 2, pp. 930–942, 2007.

[9] C. M. Bishop, "Neural Networks for Pattern Recognition", *Agricultural Engineering International the Cigr Journal of Scientific Research & Development Manuscript Pm*, vol. 12, no. 5, pp. 1235–1242, 1995.

[10] I. Kononenko, "Estimating attributes: analysis and extensions of RELIEF", *proc. European Conference on Machine Learning on Machine Learning*, 1994, pp. 171-182.

[11] X. Chen, G. Yuan, F. P. Nie, et al, "Semi-supervised Feature Selection via Rescaled Linear Regression", *proc. Twenty-Sixth International Joint Conference on Artificial Intelligence*, 2017, pp. 525-1531.

[12] L. Zhu, L. Miao, D. Zhang, "Iterative Laplacian Score for Feature Selection", *proc. Chinese Conference on Pattern Recognition*, 2012, pp. 80-87.

[13] T. Strutz, Data Fitting and Uncertainty: A Practical Introduction to Weighted Least Squares and Beyond, Vieweg and Teubner, 2010.

[14] F. P. Nie, H. Huang, X. Cai, et al, "Efficient and robust feature selection via joint $\ell_{2,1}$-norms minimization", *proc. International Conference on Neural Information Processing Systems*, 2010, pp. 1813-1821.

[15] D. Cai, X. He, J. Han, et al, "Orthogonal laplacianfaces for face recognition", *IEEE Trans Image Process*, vol. 15, no. 11, pp. 3608–3614, 2006.

[16] F. P. Nie, R. Zhang, and X. Li, "A generalized power iteration method for solving quadratic problem on the Stiefel manifold", *Science China Information Sciences*, vol. 60, no. 11, pp. 112101, 2017.

[17] F. P. Nie, S. Xiang, Y. Jia, et al, "Trace ratio criterion for feature selection", *proc. National Conference on Artificial Intelligence*, AAAI Press, 2008, pp. 671-676.

[18] F. Fleuret, "Fast binary feature selection with conditional mutual information," *Journal of Machine learning research*, vol. 5, no. 11, pp. 1531–1555, 2004.

[19] R. Zhang, F. Nie, X. Li. "Feature selection under regularized orthogonal least square regression with optimal scaling," *Neurocomputing,* vol. 273, pp. 547-553, 2018.

[20] F. Nie, R. Zhang, X. Li. "A generalized power iteration method for solving quadratic problem on the Stiefel manifold," *Science China Information Sciences*, vol. 60, no. 11, pp. 146-155, 2017.

[21] M. J. D. Powell, "A method for nonlinear constraints in minimization problems," *Optimization*, vol. 5, no. 6, pp. 283–298, 1969.

[22] D. P. Bertsekas, Constrained optimization and Lagrange multiplier methods, *Academic Press*, 1982.

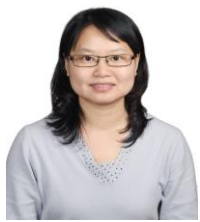

**Xia Wu** received the Ph.D. degree in basic psychology from the State Key Laboratory of Cognitive Neuroscience and Learning of BNU, in 2007. She is currently a professor at the College of Information Science and Technology of BNU, China. Her main research interests include intelligent signal processing, especially neuroimaging data analysis.

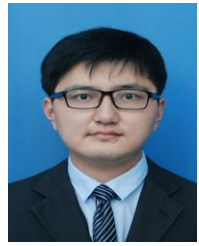

**Xueyuan Xu** is currently pursuing the Ph.D. degree in computer application technology from the College of Information Science and Technology of Beijing Normal University, China. His research interests include blind signal separation, machine learning and emtion recognition.

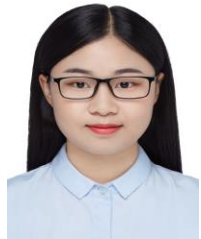

**Jianhong Liu** is a Master's degree candidate in computer application technology from the College of Information Science and Technology of Beijing Normal University, China. Her research interests focus on the machine learning and recognition of emotion based on EEG.

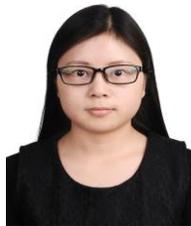

**Hailing Wang** is currently pursuing the Ph.D. degree in computer application technology from the College of Information Science and Technology of Beijing Normal University, China. Her research interests include intelligent signal processing, especially machine learning for neuroimaging data processing.

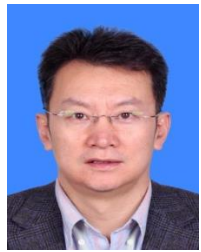

**Bin Hu** is a professor at the College of Information Science and Technology of Beijing Normal University, China; His research fields are cognitive computing, context aware computing, and pervasive computing, and has published about 100 papers in peer reviewed journals, conferences, and book chapters.

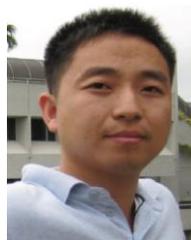

**Feiping Nie** received the Ph.D. degree in computer science from Tsinghua University, Beijing, China, in 2009. He is currently a professor at the School of Computer Science, OPTIMAL of Northwestern Polytechnical University, China. His current research interests include machine learning and its applications, such as pattern recognition, data mining, computer vision, image processing, and information retrieval.